# GFD-SSD: Gated Fusion Double SSD for Multispectral Pedestrian Detection

Yang Zheng, Izzat H. Izzat
Aptiv
{yang.zheng2, izzat.izzat}@aptiv.com

Shahrzad Ziaee
Rutgers University
shahrzad.ziaee@rutgers.edu

## Abstract

*Pedestrian detection is an essential task in autonomous driving research. In addition to typical color images, thermal images benefit the detection in dark environments. Hence, it is worthwhile to explore an integrated approach to take advantage of both color and thermal images simultaneously. In this paper, we propose a novel approach to fuse color and thermal sensors using deep neural networks (DNN). Current state-of-the-art DNN object detectors vary from two-stage to one-stage mechanisms. Two-stage detectors, like Faster-RCNN, achieve higher accuracy, while one-stage detectors such as Single Shot Detector (SSD) demonstrate faster performance. To balance the trade-off, especially in the consideration of autonomous driving applications, we investigate a fusion strategy to combine two SSDs on color and thermal inputs. Traditional fusion methods stack selected features from each channel and adjust their weights. In this paper, we propose two variations of novel Gated Fusion Units (GFU), that learn the combination of feature maps generated by the two SSD middle layers. Leveraging GFUs for the entire feature pyramid structure, we propose several mixed versions of both stack fusion and gated fusion. Experiments are conducted on the KAIST multispectral pedestrian detection dataset. Our Gated Fusion Double SSD (GFD-SSD) outperforms the stacked fusion and achieves the lowest miss rate in the benchmark, at an inference speed that is two times faster than Faster-RCNN based fusion networks.*

1. Introduction

Pedestrian detection is one of the most essential tasks in the area of autonomous driving research and development. It is also the key barrier to deploy automated vehicles in urban areas. Despite the fact that this challenging problem has been intensively progressed among artificial intelligence studies [1, 2, 3, 4] in recent years, there is still a huge gap towards a satisfactory especially in critical environments such as weak illumination, far distance, and occlusion of pedestrians. Typical computer vision approaches take color images as the input. In addition, near infrared (0.75~1.3µm) cameras or long-wavelength infrared (7.5~13µm, i.e., thermal) cameras provide complementary information beyond the visual spectrum. Since long-wavelength infrared images are more robust to the interferences produced by lighting conditions, headlights and traffic signals, pedestrians are more visible in thermal cameras and hence can be reliably detected. With the availability of KAIST multispectral pedestrian detection dataset [5], many state-of-the-art studies demonstrated that detection performance can be improved by combining a well-aligned pair of color and thermal sensors [6, 7, 8, 9, 10].

Modern DNN-based object detectors are generally categorized into two-stage and one-stage detectors. Two-stage detectors, such as Faster-RCNN [11] and R-FCN [12], are based on proposal-driven mechanisms. The first stage generates a sparse set of object candidate locations through a Region Proposal Network (RPN) on the feature map, and the second stage classifies each candidate as foreground or background while jointly performing localization regression. On the contrary, one-stage approaches like SSD [13] and YOLO [14] eliminate proposal generation and subsequent feature resampling, while encapsulating all computation in a single network. SSD adopts its backbone network (e.g., VGG16 [15]) features in a pyramid structure to detect objects with different scales. Features are formulated in a pyramid structure which contains high-level semantic information as well as low-level localization context. This is beneficial to detect small objects from shallow layers and recognize large objects from deeper layers. A comprehensive study in [16] compares the trade-off between accuracy and speed among several modern object detectors, which demonstrates an overall conclusion that two-stage detectors achieve higher accuracy while one-stage detectors perform better in speed. In consideration of autonomous driving applications which require real-time accurate detection, we develop our network architecture based on a one-stage detector and improve its accuracy to be comparable or superior to two-stage detectors.

For the multispectral pedestrian detection with the fusion of color and thermal modalities, all of the studies [6, 7, 8, 9] on the KAIST benchmark follow the two-stage approach. They employ Faster-RCNN as the main architecture and VGG16 as the backbone feature extractor. Although satisfactory accuracies were achieved, the computation times are not reported in any of studies. Their fusion strategies vary from early fusion, halfway fusion, and late fusion, depending on which convolutional layers in RPN are used to combine the color and thermal subnetworks. The fusion method itself is based on concatenating selected feature maps and applying a subsequent Network-In-Network (NIN) [17] to reduce the feature map dimension. SSD differs from Faster-RCNN in that its feature pyramid is constructed by multiple feature maps ranging from early to late stage convolutional layers. A typical fusion approach is to concatenate all feature maps from color and thermal modalities, which will double the number of anchors for the final classification layer. We call this stacked fusion. In this paper, we propose a novel gated fusion structure which utilizes the Gated Fusion Unit (GFU). The GFU works as an intermediate bridge between the color SSD and thermal SSD, which takes the input from two feature maps and outputs a joint feature map without changing the size. The operation of GFU is based on a gating mechanism endeavoring to keep the information delivered by reliable features while mitigating the effect of degraded features. The design of GFU is motivated by [18], and we further modify and derive two versions, GFU_v1 and GFU_v2, from it. The major difference between the two versions is that GFU_v1 applies a convolution kernel on the concatenated feature, while GFU_v2 applies two kernels on two individual features. Leveraging GFUs on the feature pyramid, the entire double-SSD architecture varies from complete gated fusion to mixed fusion. The mixed fusion is to apply GFUs on a subset of feature maps while maintaining others stacked. We further design the mixed fusion into four different versions, Mixed_Early, Mixed_Late, Mixed_Even, and Mixed_Odd, depending on which layers are selected to use the GFU. Evaluating on the KAIST dataset, our Mixed_Even network is shown to achieve the competitive state-of-the-art accuracy, while performing twice as fast as Faster-RCNN based two-stage fusion architectures. Note that the proposed GFUs and fusion architectures are not limited to color and thermal images but could be readily applied to any CNN-based sensor fusion.

The major contributions of this paper are summarized as follows:
(1) We design two versions of GFU to operate the combination of two convolutional feature maps. The joint output learns weights to adjust the contribution of each feature map.
(2) We propose one gated fusion and four mixed fusion architectures on the feature pyramid to integrate two SSDs on different modalities. The entire network is an end-to-end one-stage object detector.
(3) With quantitative and qualitative experiments on the KAIST multispectral pedestrian detection dataset, we demonstrate that our Gated Fusion Double SSD (GFD-SSD) can achieve the state-of-the-art accuracy while performing 2× faster than two-stage detectors.

2. Related Work

2.1. Multispectral Pedestrian Detection

KAIST multispectral pedestrian detection benchmark [5], to the best of our knowledge, is the largest public dataset that provides calibrated, annotated color and thermal images. Image pairs from video streams were synchronized at a rate of 20FPS and calibrated into a uniform resolution of 640×512 pixels. In total, KAIST contains 95.3k pairs of color-thermal images, separated into a training set of 50.2k images with 41.5k pedestrians and a testing set of 45.1k images with 44.7k pedestrians.

The KAIST benchmark baseline utilized a multispectral Aggregated Channel Features (ACF) detector [19] with the combination of thermal intensity and Histogram of Oriented Gradients (HOG) features, and obtained 50.48% log-average miss rate (logMR, measured within the false positive per image range [$10^{-2}$, $10^0$]) for the overall performance in both day and night times. However, this classical approach is based on hand-crafted features without benefiting from the success of DNNs. In [6], a Faster-RCNN architecture was investigated with early-fusion and late-fusion strategies to combine the information from two subnetworks. In addition, [7] suggested a halfway-fusion model to better carry low-level fine visual details and high-level semantic abstraction. [9] further improved the halfway-fusion accuracy by adding a Boost Decision Tree (BDT) [20] classifier on top of the fused RPN to reduce potential false-positive detections. Due to the correlation of pedestrian detection confidence with illumination condition in both cases of color and thermal images, [10] designed an Illumination-aware Faster-RCNN (IAF R-CNN) to adjust the final confidence score over the illumination value. However, the two-stage detector Faster-RCNN achieves better accuracy with the cost of greater computation complexity, which is not appropriate for real-time autonomous driving applications. Moreover, Faster-RCNN has a limited performance for small, low-resolution objects [21], such as far-distance pedestrians. Therefore, SSD based object detectors are considered.

## 2.2. Variations on Single Shot Detectors

SSD adopts a hierarchical pyramid formation of selected feature maps. Shallow layers with small receptive fields and high resolutions focus on locating small objects, while deep layers with large receptive fields are comparatively suitable for detecting large objects. This ensures its capability to handle multiple scale objects in a fast one-stage detector. DSSD [22] extends SSD by adding deconvolutional layers onto higher level feature maps to increase their resolution and replace the original VGG16 backbone network with ResNet101 [23], thus creating an encoder-decoder architecture. Instead of directly using the backbone network feature maps, FSSD [24] concatenates different scale layers and generates a series of new feature pyramids. RetinaNet [25] uses a feature pyramid network backbone on top of a feedforward ResNet architecture and introduces a focal loss function to eliminate the accuracy gap between one-stage and two-stage detectors. These variations are based on one single SSD model without fusion of different modalities. DGFN [18] considers two separate CNNs on RGB and Lidar images, and connects the two networks via a gated fusion unit on intermediate layers. This motivates us to design updated versions of GFU, and to deploy more structural variations for pedestrian detection on color and thermal images.

## 3. Gated Fusion Double SSD

### 3.1. Overall Gated Fusion Architecture

In an original SSD with a 300×300 input image size, the feature pyramid is constructed by the layer outputs at conv4_3, conv7 (FC7), conv8_2, conv9_2, conv10_2, and conv11_2. For two SSDs, a simple stack fusion strategy (shown in Figure 1-a) is to concatenate all layers from two pyramids, which will double the final feature map dimensions and the associated number of anchors. Figure 1-b shows the gated fusion architecture, which inputs the same layer feature maps from two SSDs into a GFU and collects all GFU outputs into the detector feature pyramid. The GFU output is a joint combination of two input feature maps, and its dimension equals the dimension of one input. Therefore, the number of anchors from GFD-SSD maintain the same with a single SSD.

### 3.2. Gated Fusion Unit (GFU)

GFU is the key component that adjusts the feature map combination between two modalities. We propose two versions of GFU design as shown in Figure 2. In the figure, $F_C$ and $F_T$ are the corresponding input feature maps from color and thermal SSDs, and $F_G$ is the concatenation of $F_C$ and $F_T$. In the first GFU design (GFU_v1, shown in Fig 2-a), we apply two 3×3 kernels ($w_C$, $b_C$) and ($w_T$, $b_T$) on $F_G$ separately, followed by a Rectified Linear Unit (ReLU) activation function on each path. $A_C$ and $A_T$ denote the activation outputs. We then perform an element-wise summation on $A_C$ and $A_T$ with $F_C$ and $F_T$ respectively. $F_F$ denotes the concatenation of the two summations, which is then passed to a 1×1 kernel ($w_J$, $b_J$) to generate the joint feature output $F_J$. GFU_v1 is similar to the design in [18], but differs in that we use ReLU instead of sigmoid activation function to generate $A_C$ and $A_T$, and then perform element-wise summation instead of the product operation on $F_C$ and $F_T$. Since the output range of sigmoid function is (0, 1), multiplying a (0, 1) value on the input feature map is a complicated form of weighting mechanism. This could be equivalently accomplished by top detection layers outside GFU. We intended to break the (0, 1) constrain by replacing the sigmoid with ReLU which has an output range (0, +∞), and replacing the following product operation with summation. The two 3×3 kernels ($w_C$, $b_C$) and ($w_T$, $b_T$) in GFU_v1 are applied on the concatenated feature map $F_G$, whereas in the second version of GFU (GFU_v2, shown in Fig 2-b), these kernels are applied on $F_C$ and $F_T$ individually. The intuition for this alternative design is that GFU_v1 keeps the original inputs and learns operational adjustments from their combination, whereas GFU_v2 learns adjustments directly from the original input.

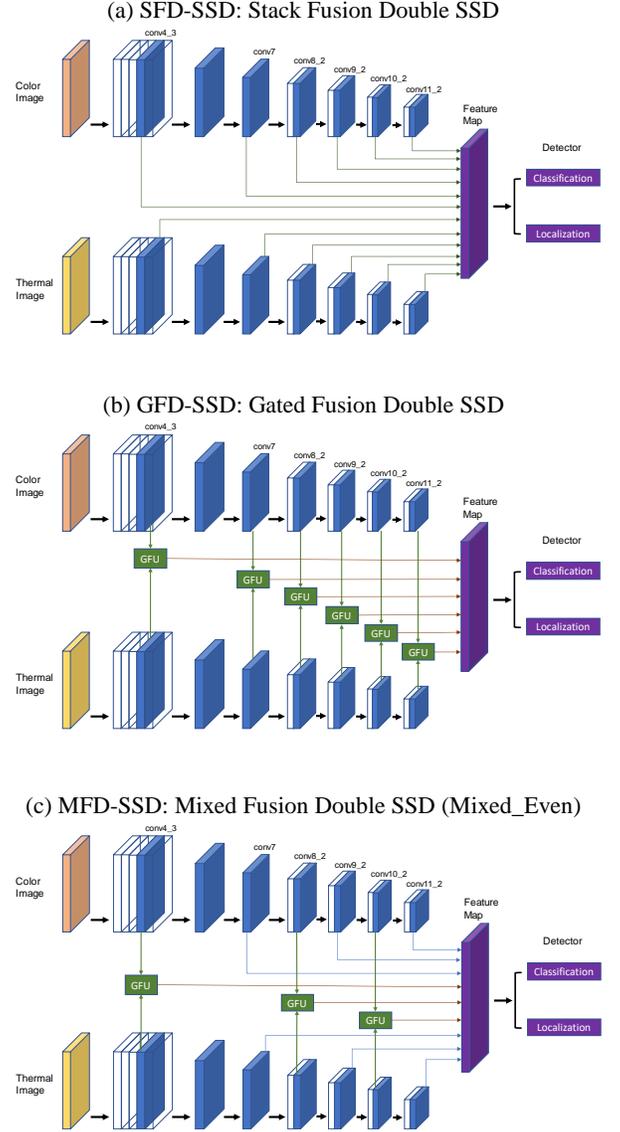

Figure 1: Overall architecture of (a) stacked fusion, (b) gated fusion, and (c) mixed fusion structures of double SSD300.

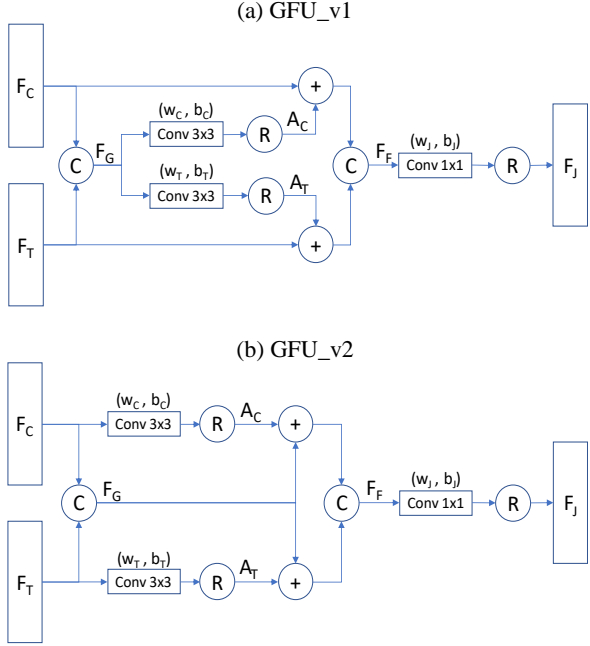

Figure 2: Two versions of GFU design. $F_C$ and $F_T$ denote input color and thermal features respectively, $F_J$ denotes the joint output from GFU. ⓡ denotes ReLU activation function, ⓒ denotes concatenation, and ⊕ denotes element-wise summation.

| Model | Feature Pyramid | Num. of Anchors |
|---|---|---|
| Single SSD | conv4_3, conv7, conv8_2, conv9_2, conv10_2, conv11_2 | 8,732 |
| Stack Fusion | conv4_3_C, conv4_3_T, conv7_C, conv7_T, conv8_2_C, conv8_2_T, conv9_2_C, conv9_2_T, conv10_2_C, conv10_2_T, conv11_2_C, conv11_2_T | 17,464 |
| Gated Fusion | conv4_3_G, conv7_G, conv8_2_G, conv9_2_G, conv10_2_G, conv11_2_G | 8,732 |
| Mixed_Even | conv4_3_G, conv7_C, conv7_T, conv8_2_G, conv9_2_C, conv9_2_T, conv10_2_G, conv11_2_C, conv11_2_T | 11,052 |
| Mixed_Odd | conv4_3_C, conv4_3_T, conv7_G, conv8_2_C, conv8_2_T, conv9_2_G, conv10_2_C, conv10_2_T, conv11_2_G | 15,144 |
| Mixed_Early | conv4_3_G, conv7_G, conv8_2_G, conv9_2_C, conv9_2_T, conv10_2_C, conv10_2_T, conv11_2_C, conv11_2_T | 8,922 |
| Mixed_Late | conv4_3_C, conv4_3_T, conv7_C, conv7_T, conv8_2_C, conv8_2_T, conv9_2_G, conv10_2_G, conv11_2_G | 17,274 |

Table 1: Summary of fusion structures. The original SSD is based on a VGG16 backbone and image size 300×300. Feature map notation suffixes "_C", "_T", and "_G" represent convolutional layer outputs from color SSD, thermal SSD, and GFU, respectively.

The operations of the GFUs are summarized in the following equations.

For GFU_v1,
$$\begin{aligned} F_G &= F_C \odot F_T \\ A_C &= \mathrm{Re}\,LU(w_C * F_G + b_C) \\ A_T &= \mathrm{Re}\,LU(w_T * F_G + b_T) \\ F_F &= (F_C \oplus A_C) \odot (F_T \oplus A_T) \\ F_J &= \mathrm{Re}\,LU(w_J * F_F + b_J) \end{aligned} \quad (1)$$

For GFU_v2,
$$\begin{aligned} F_G &= F_C \odot F_T \\ A_C &= \mathrm{Re}\,LU(w_C * F_C + b_C) \\ A_T &= \mathrm{Re}\,LU(w_T * F_T + b_T) \\ F_F &= (F_G \oplus A_C) \odot (F_G \oplus A_T) \\ F_J &= \mathrm{Re}\,LU(w_J * F_F + b_J) \end{aligned} \quad (2)$$

where
- $\odot$: concatenation
- $\oplus$: element-wise summation
- $F_C$, $F_T$, $F_G$, $F_F$, $F_J$: feature maps
- $A_C$, $A_T$: ReLU activation outputs
- $w_C$, $w_T$, $w_J$: kernel weights
- $b_C$, $b_T$, $b_J$: kernel biases

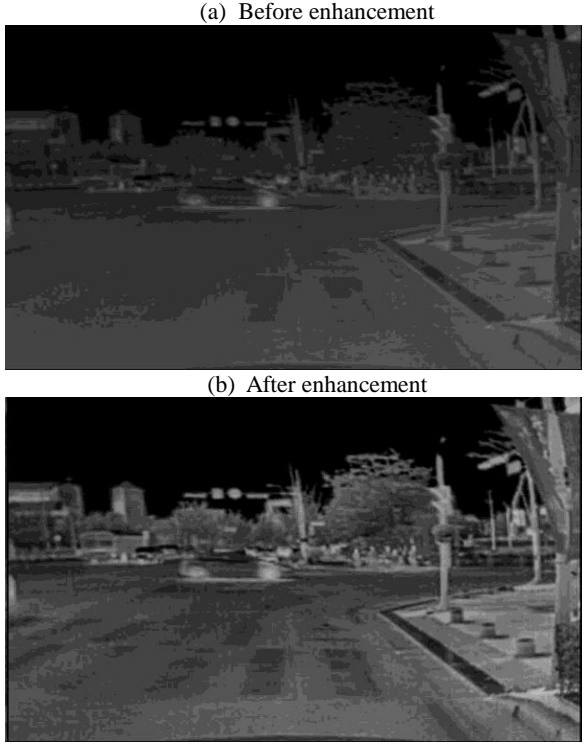

Figure 3: Comparison of example thermal images (a) before and (b) after enhancement.

### 3.3. Mixed Fusion Variations

The gated fusion structure shown in Figure 1-b applies GFUs on all feature maps within the VGG16 backbone pyramid. Alternatively, the mixed fusion selects a subset of feature maps for gated fusion and keep the remaining features in stack fusion. We design four variations of mixed fusion named Mixed_Even, Mixed_Odd, Mixed_Early, and Mixed_Late, depending on which layers are selected to use the GFUs. For example, Figure 1-c illustrates the structure of Mixed_Even, which applies GFUs on conv4_3, conv8_2, and conv10_2. On the contrary, Mixed_Odd keeps these layers stacked and gates the others. Mixed_Early applies GFUs on shallower feature maps while Mixed_Late applies GFUs on deeper layers. Since SSD uses pre-set anchors which are associated with the feature pyramid, mixed fusion variations can result in different number of anchors. In Table 1, we summarize the number of anchors based on an original SSD with VGG16 backbone and image size 300×300, the proposed fusion architectures and their associated number of anchors. A larger number of anchors provides more potential object candidates, but also increases the required computational resources. For the input image size 512×512, the single SSD has an extra conv12_2 layer in its feature pyramid. Therefore, we let conv12_2 be gated in Mixed_Even and Mixed_Late and stacked in Mixed_Odd and Mixed_Early.

### 3.4. Loss Function

The total loss function $L_{total}$ is defined as a weighted summation of three components: classification loss $L_{cls}$, localization loss $L_{loc}$, and $L2$ regularization loss with weight decay. Since we fuse two SSDs into one architecture, the number of model parameters is increased. Therefore we reduce the impact of $L2$ regularization loss by decreasing its weight $\gamma$. For our task on the KAIST dataset, labels contain objects of types "person" (single pedestrian), "people" (a group of pedestrians that are hard to separate), "cyclist", and "person?" (person-like objects, e.g., statue). Generally speaking, these are all high-level conceptual pedestrians and deserve equal attention in autonomous driving. Since localization issue is more critical than classification for autonomous driving, we use a greater value of localization loss weight $\beta$ compared to the value of classification loss weight $\alpha$. The final weights are selected in a way that they satisfy the following relative proportion: $\alpha : \beta : \gamma = 5:10:1$. The total loss used in our work is defined by Equation (3).

$$L_{total} = \alpha L_{cls} + \beta L_{loc} + \gamma L2 \qquad (3)$$

To handle the imbalance between positive and negative anchors, we adopt the idea of Online Hard Example Mining (OHEM) [26] which was originally applied to Faster-RCNN. We keep the foreground-to-background ratio in each mini batch to a minimum target of 1:3 by selecting the top K hard negative anchors. The classification loss is then computed in an inverse class weighting form, where $N^+$ and $N^-$ denote the number of selected positive/negative anchors, and $L^+$ and $L^-$ denote their softmax cross-entropy loss respectively. The localization loss is computed as a summary of smooth L1 loss. Note that although Focal Loss is claimed to be more effective in [25], in our experiments, we found out that it yields to slightly better precision but worse recall. Since KAIST benchmark results are measured by log-average miss rates, we decide not to apply Focal Loss in our models. The classification loss used in our experiments is presented in Equation (4).

$$L_{cls} = \frac{N^-}{N^+ + N^-} L^+ + \frac{N^+}{N^+ + N^-} L^- \qquad (4)$$

### 4. Experiments

#### 4.1. Pre-processing

**Thermal Enhancement:** In the KAIST dataset, raw thermal images have a relatively lower resolution (320×256) than color images (640×480), which makes it difficult to fully utilize the information in recognition tasks. Although the color-thermal pairs have been virtually aligned into a uniform size 640×512, such an interpolation does not improve the thermal observation. Thermal enhancement [27] is therefore suggested. We employ a contrast-limited adaptive histogram equalization technique to adjust image intensities and enhance contrast. Figure 3 displays an example to compare the difference before and after thermal enhancement.

**Data Augmentation:** As analyzed in [13], data augmentation is crucial for SSD. For the model training, we employ a series of transformations on color images, including changes in brightness, contrast, HUE, saturation, and RGB channels ordering. We also employ horizontal flip and resize transformations on both color and thermal images. Each transformation is randomly applied with a 50% probability.

**Transfer Learning:** For data-driven deep neural network approaches, more data is always helpful. The largest real-scenario pedestrian detection corpus is the Caltech dataset [28], but it only contains color images in the daytime. In addition to KAIST, a recent CVC14 dataset [29] provides a relatively small-scale multi-modal pedestrian detection data in both day and night environments. However, their visual images are in grayscale, and their visual-thermal pairs are not well aligned, thus makes it error-prone. More recently, a thermal camera manufacturer (FLIR) released a starter dataset on their website[1], which includes over 10K pairs of color and thermal images with bounding box annotations. However, in this dataset, color and thermal cameras have different focal length and resolutions, making it necessary to perform additional pre-process to align the pairs. In our experiment, we initialize our model with a pre-trained VGG16 backbone on ImageNet, fine-tune the entire model using the FLIR dataset, and finally re-train it using KAIST.

### 4.2. Results on KAIST dataset

**SSD300 vs. SSD512:** A single SSD model with larger input image size (SSD512) has been shown in [13] to achieve higher accuracy than the one with smaller input image size (SSD300) on PASCAL VOC and COCO datasets. This can be explained by the fact that larger input size leads to higher resolutions on feature maps and a greater number of anchor locations and aspect ratios, which are essential to help to detect small objects. In our experiments on the KAIST dataset, we also observe from Table 2 and Table 3 that SSD512 has an overall superior performance to SSD300, for both single modality SSDs and double fusion SSDs. Figure 4 plots the miss rates over the false positives per image (FPPI) within the range $[10^{-2}, 10^0]$. It can be observed that SSD512 has a greater magnitude range of miss rates than SSD300. With the increase of FPPI, SSD300 is easier to converge to its minimum miss rate than SSD512.

[1] www.flir.com/adasdataset

| Model | Inputs | Fusion | logMR (%) |
|---|---|---|---|
| Single SSD300 | Color | - | 34.69 |
| | Thermal | - | 41.79 |
| Double SSD300 | Color + Thermal | Stack | 29.99 |
| | Color + Thermal | GFU_v1 | 30.42 |
| | Color + Thermal | GFU_v2 | 30.51 |
| Single SSD512 | Color | - | 32.81 |
| | Thermal | - | 39.47 |
| Double SSD512 | Color + Thermal | Stack | 30.29 |
| | Color + Thermal | GFU_v1 | 28.84 |
| | Color + Thermal | GFU_v2 | **28.10** |

Table 2: Result summary on KAIST pedestrian detection, for comparisons of SSD300 vs. SSD512, and stack fusion vs. gated fusion v1 vs. gated fusion v2.

| Model | Mix Type | logMR (%) | | |
|---|---|---|---|---|
| | | Overall | Day | Night |
| Double SSD300 + GFU_v2 | Mixed_Even | 28.06 | 26.07 | **29.22** |
| | Mixed_Odd | 29.69 | 28.19 | 30.57 |
| | Mixed_Early | **28.00** | **25.80** | 30.03 |
| | Mixed_Late | 28.11 | 26.16 | 29.33 |
| Double SSD512 + GFU_v2 | Mixed_Even | 27.41 | 26.67 | 28.66 |
| | Mixed_Odd | 29.67 | 30.83 | 29.24 |
| | Mixed_Early | **27.17** | **25.28** | 27.49 |
| | Mixed_Late | 27.75 | 28.04 | **26.74** |

Table 3: Result summary on KAIST pedestrian detection, for comparisons of variant mixed fusion using double SSD and GFU_v2, and the difference in day and night times.

| | Model | logMR (%) | Infer. time (sec./image) |
|---|---|---|---|
| Ours | **Double SSD300 + Stack Fusion** | **29.99** | **0.0441** |
| | **Double SSD300 + GFU_v2** | **30.51** | **0.0585** |
| | **Double SSD300 + Mixed Early** | **28.00** | **0.0512** |
| Reproduction | Single SSD300 + Color Input | 34.69 | 0.0300 |
| | SSD + MobileNet (pixel-fusion) | 71.98 | 0.0124 |
| | Faster-RCNN + Inception (pixel-fusion) | 51.20 | 0.3426 |
| | Faster-RCNN + Resnet (end-to-end mid-fusion) | 30.57 | 0.1026 |
| KAIST Benchmark | FusionRPN + BDT [9] | 29.83 | - |
| | IAF R-CNN [10] | 33.19 | - |
| | Halfway Fusion [7] | 36.22 | - |
| | Late Fusion CNN [6] | 43.80 | - |
| | CMT-CNN [8] | 49.55 | - |
| | Baseline, ACF+T+THOG [5] | 54.40 | - |

Table 4: Result summary on KAIST pedestrian detection, for the miss rate accuracy and speed performance against state-of-the-art models.

**Stack vs. GFU_v1 vs. GFU_v2:** For the fusion of

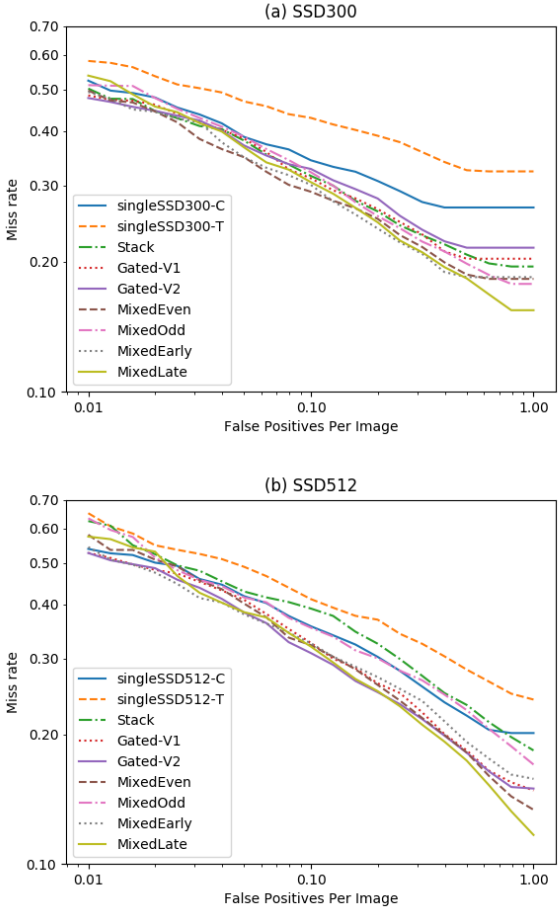

Figure 4: Plots of miss rates over the false positives per image (FPPI) within the range $[10^{-2}, 10^0]$. Comparison of detection results with single SSD, stack fusion, gated fusion, and mixed fusion of double SSDs.

double SSDs, Table 2 compares the difference between stack fusion and gated fusions (GFU_v1 and GFU_v2). For SSD300, the difference is not significant, and the stack fusion shows an even better performance than the gated fusions. For SSD512, however, the gated fusions have noticeably better performances than the stack fusion, and the GFU_v2 achieves the lowest logMR: 28.10%. Since SSD300 decreases the input image resolution, we infer our GFUs are more effective for higher resolution images or feature maps.

**Mixed Fusion:** We further investigate four types of mixed fusion of double SSD300 and SSD512, using GFU_v2 as the connection unit. As shown in Table 3, we examine their overall accuracies in the day and night environments. The Mixed_Early achieves the lowest overall logMR for both SSD300 (28.00%) and SSD512 (27.17%). Early level convolutional layers have higher resolutions with more contextual details, while late level layers are more conceptually abstract with lower resolutions. We further claim that the GFUs are more effective in adjusting the combination of low-level and high-level features. The best detection result in the daytime is also achieved by Mixed_Early, but the night time performance varies from case to case.

**Detection Accuracy and Inference Time:** Table 4 compares our work against other state-of-the-art studies. According to the most recent KAIST benchmark results, the best pedestrian detection accuracy (i.e., the lowest log average miss rate) has been achieved at 29.83% logMR by [9]. Unfortunately, none of the benchmark studies reported their speed performances. This could be due to the fact that their fusion strategies are divided into two separate stages, which makes it difficult to measure the inference time. They first take mid-level feature outputs from two modality subnetworks and then concatenate them as the input of the second stage for fine-tuning. Therefore, the weights of lower-layer subnetworks are fixed and only the upper-layers after fusion are trained. On the contrary, our work is end-to-end. For this ablation study, we selected SSD300 rather than SSD512 since it results in remarkably faster inference while not losing too much accuracy. The Mixed_Early fusion of double SSD300 achieves the lowest logMR at 28.00% while spending 0.0512 seconds per image on a single Nvidia GeForce GTX 1080 Ti GPU card. For reference, we follow the "meta-architecture + feature extractor" configurations in [16] and reproduce experiments using the KAIST dataset. Specifically, we investigate pixel-fusions (append thermal as an additional channel to RGB) on SSD with MobileNet [30] and Faster-RCNN with Inception [31], as well as an end-to-end mid-level feature-fusion on Faster-RCNN with ResNet [23]. The "SSD + MobileNet" configuration is designed for light-weighted implementation. It is the fastest model but also the worst in accuracy, which is unacceptable in the autonomous driving domain. The "Faster-RCNN + ResNet" model obtains a comparable 30.57% logMR with other KAIST benchmark studies, and its inference time is 0.1026 second/image. Our double SSD fusion models are approximately twice as fast.

**Examples:** Figure 5 presents four examples for results visualization, two in the daytime and two in the night time. Example Day_1 contains a person riding a motorcycle. According to our experiment, the single SSD with color input made a mistake by predicting two separate labels (person and cyclist), while the single thermal SSD, GFD-SSD and Mixed_Early fusion models correctly predicted one label. The annotation of example Day_2 contains a mixture of several single "Person" labels and a "People" label for a clustered group. The single color SSD resulted in more false negatives, while the single thermal SSD resulted in more false positives. The GFD-SSD and Mixed_Early detected very similar results by balancing the two single SSDs. Indeed, the annotated labels for this example are not accurate enough. In the example Night_1, the single color SSD missed one pedestrian, while the single thermal SSD detected a group of people on the left side, which is difficult to detect even by humans. Both the GFD-SSD and Mixed_Early models detected two pedestrians on the right side, but Mixed_Early obtained better localization accuracy. In the example Night_2, again, the annotated labels are not that accurate to cover all the pedestrians. The Mixed_Early achieved better results compared to the other three, and much closer to the ground truth.

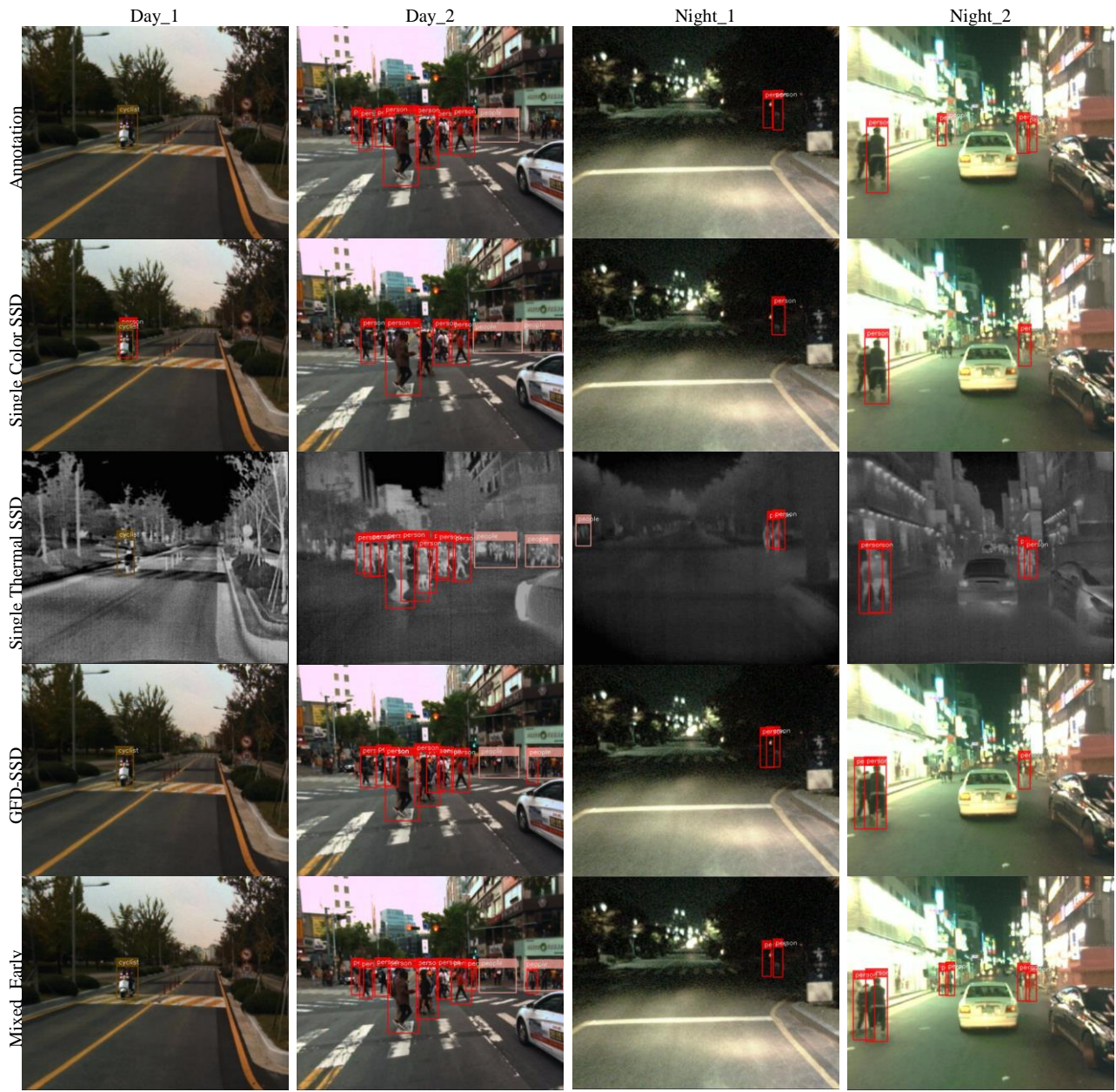

Figure 5: Examples using single and double SSDs for KAIST pedestrian detection. Four columns list 2 examples in the daytime and 2 in the night. Row 1 is for the annotated label; Row 2 is for the result using a single SSD with color inputs; Row 3 is for a single SSD with thermal inputs; Row 4 is for the gated fusion double SSD (GFD-SSD) using GFU_v2; Row 5 is for the Mixed_Early fusion.

5. Conclusion

In this paper, we focused on the task of multispectral pedestrian detection using color and thermal image pairs.

We designed two versions of gated fusion units to operate the connection between two modality SSD feature layers. Additionally, we proposed four variant mixed fusion architectures. Experiments conducted on KAIST dataset show that the combination of double SSDs outperforms single SSDs, and our gated and mixed fusions are superior to the traditional stack fusion. Comparing with the state-of-the-art Faster-RCNN based two-stage detection fusion models, our one-stage GFD-SSD and its mixed variations achieved both competitive accuracy and better inference runtime.

As a data-driven deep learning model, one potential improvement is to add more data for transfer learning. Datasets like Caltech and CVC contain either color or thermal single modality images. Although these datasets are not available in pairs, they may be useful if we freeze one subnetwork SSD at a time and finetune the other. More experiments will be conducted in the continued work. In addition to color and thermal images, the fusion strategies presented in this paper are also beneficial to other types of input, like the 3D depth information captured from laser sensors [32] and static street masks generated from map data [33].